# Streaming Batch Eigenupdates for Hardware Neuromorphic Networks

**Brian D. Hoskins[1]\*, Matthew W. Daniels[1], Siyuan Huang[2], Advait Madhavan[1,3], Gina C. Adam[2], Nikolai Zhitenev[1], Jabez J. McClelland[1], Mark D. Stiles[1]**

[1]Physical Measurement Laboratory, National Institute of Standards and Technology, Gaithersburg, MD, USA

[2]Electrical and Computer Engineering, George Washington University, Washington, DC, USA

[3]Institute for Research in Electronics and Applied Physics, University of Maryland, College Park, MD, USA

**\* Correspondence:**
Brian D. Hoskins
brian.hoskins@nist.gov

## 1    Abstract

Neuromorphic networks based on nanodevices, such as metal oxide memristors, phase change memories, and flash memory cells, have generated considerable interest for their increased energy efficiency and density in comparison to graphics processing units (GPUs) and central processing units (CPUs). Though immense acceleration of the training process can be achieved by leveraging the fact that the time complexity of training does not scale with the network size, it is limited by the space complexity of stochastic gradient descent, which grows quadratically. The main objective of this work is to reduce this space complexity by using low-rank approximations of stochastic gradient descent. This low spatial complexity combined with streaming methods allows for significant reductions in memory and compute overhead, opening the doors for improvements in area, time and energy efficiency of training. We refer to this algorithm and architecture to implement it as the streaming batch eigenupdate (SBE) approach.

## 2    Introduction

Deep neural networks (DNNs) have grown increasingly popular over the years in a wide range of fields from image recognition to natural language processing. These systems have enormous computational overhead, particularly on multiply and accumulate (MAC) operations, and specialized hardware has been made to accelerate these tasks. As the networks are themselves tolerant to noise and low precision computing (4-bit and below), theoretical and experimental investigations have shown that analog implementations of DNNs using Ohm's and Kirchoff's laws to perform MAC operations can vastly accelerate the training and reduce the energy of inference by orders of magnitude.

Investigations regarding an appropriate nanodevice suitable for analog inference have focused on different families of 2-terminal memory devices (memristors, resistive random-access memory (ReRAM), phase change memories (PCM), etc.) as well as 3 terminal devices (flash memory, lithium insertion) (Haensch, Gokmen, and Puri 2019). These devices have the desirable properties of analog tunability, high endurance, and long-term memory needed for use in embedded inference applications. Applications based on these devices perform well when used for inference and have been well studied, with intermediate scale systems having been being built by integrating devices into crossbar arrays (Adam et al. 2017; Prezioso et al. 2015; Chakrabarti et al. 2017; Z. Wang et al. 2018).

Though most of the effort has been focused on building inference engines, more recent work has begun to address difficulties in training such nanodevice arrays (Adam 2018; Ambrogio et al. 2018). In crossbar architectures, there are two approaches to updating the weights. The first, which fits well with weights computed in software, is to sequentially update each weight separately. The other, called an outer product update, is to update all the weights simultaneously with two vectors of voltages or voltage pulses. This latter approach is limited in the type of updates that can be applied, but its speed and energy advantage essentially preclude the use of the former in practical applications. The goal of the work presented here is to develop a technique based on outer product updates that approaches the training fidelity available for algorithms based on sequential weight updates, which are often employed in software-based platforms.

A core advantage of such a system is that it's been observed to have $O(1)$ complexity for the most critical operations, inference and update. For a suitably parallelized architecture, the number of clock cycles needed for these operations is independent of the size of the memory arrays in each layer (T. Gokmen and Vlasov 2016). For inference, this is not a problem, but for matrix updates, this limits the training algorithm to stochastic gradient descent (SGD), since this is the only algorithm which uses rank 1 outer product updates alone. This approach does not allow independent updates of each element; therefore, a complete high rank update of a whole crossbar would require a series of these outer product updates. Though SGD is a powerful method for training, other methods, employed in software, such as momentum, Adagrad, or, most simply, batch update can sometimes be superior. However, these require additional memory overhead or explicit look-a-head updates of the memory (T. Gokmen, Rasch, and Haensch 2018).

Mini-batch gradient descent (MBGD), as the simplest possible modification of SGD, is of extreme interest, particularly in the case of nanodevice arrays. It's been suggested that it can increase tolerance with respect to device nonidealities, as well as be employed to minimize the number of device updates, which can be a problem in systems with low endurance or high energy to program (Kataeva et al. 2015). In PCM arrays, minimizing the number of updates is critical to preventing a hard reset when the device reaches its natural conductance limit (Boybat et al. 2017). Additionally, in cases where the energy of inference is significantly less than the energy of update, reducing the number of updates could result in a substantial decrease in the energy required to train the network, even if it occurs at the expense of training time.

Here, we propose a new approach to batch update, one in which a batch is broken up into a smaller number of principal components which are subsequently used to train the network. This approach potentially yields some of the critical benefits of batch update but requires substantially less overhead and has a significantly lower computational cost.

## 3    Materials and Methods

### 3.1    Proposed methods for training and new algorithm

The key idea behind our alternative approach is to estimate the most representative single outer product update for the batch. Not only is this approach fast, it also minimizes the amount of information that needs to be stored to make each update. We consider then, an arbitrary network layer being trained on batch $i$ with an $a \times b$ weight matrix $\mathbf{w}^i$. The layer receives $j$ activations $\mathbf{x}^{i,j}$ of dimension $b$ and backpropagated errors $\boldsymbol{\delta}^{i,j}$ of dimension $a$ per batch. In the ideal case, we would like the network to update according to

$$\mathbf{w}^{i+1} = \mathbf{w}^i + \Delta \hat{\mathbf{w}}^i,$$

where the batch average update $\Delta \hat{\mathbf{w}}^i$ is a sum of outer products,

$$\Delta \hat{\mathbf{w}}^i = -\frac{\eta}{n} \sum_{j=1}^{n} \boldsymbol{\delta}^{i,j} [\mathbf{x}^{i,j}]^{\,T}.$$

Each term in this sum is the gradient of the loss function of that input $\mathbf{x}^{i,j}$, which is a rank 1 matrix. The sum of the gradients, $\Delta \hat{\mathbf{w}}^i$, is the gradient of the batch loss function and is in general a rank $n$ matrix. Performing such an update with conventional SGD will require $n$ outer product operations. An important observation here is that the outer product operation itself is a rank 1 operation, and hence an efficient alternative would entail using a low rank matrix approximation of $\Delta \hat{w}^i$ to reduce the total number of updates performed. More specifically, we perform $k < n$ outer product updates where $k$ is the number of significant singular values of the true gradient $\Delta \mathbf{w}^i$.

Performing the singular value decomposition (SVD) of $\Delta \mathbf{w}^i$ entails significant memory overhead and computational cost, and has been studied extensively in the computer science literature. One solution involves employing unsupervised techniques such as streaming principal component analysis (PCA) to extract the $k$ most significant singular vectors of a streaming covariance matrix (Balsubramani, Dasgupta, and Freund 2013; C. L. Li, Lin, and Lu 2016; I. Mitliagkas, Caramanis, and Jain 2013; P. Yang, Hsieh, and Wang 2018). Oja's original algorithm for PCA was developed to describe the evolution of neural network weights(E. Oja 1982). By applying his formalism here on the weight *updates*, we can extract, on the fly, a set of $k$ most representative vectors, of the original rank $n$ update. This allows us to perform memory limited batch updates with $k(a + b)$ additional memory units instead of $a \times b$ as used in previous studies. This amounts to using a separate unsupervised neural network to train the network of interest, but this network trains on the batch gradient and only needs very short-term memory as the gradient is constantly changing. Such short term memory arrays are already entering into use(Ambrogio et al. 2018).

If we consider the special case of $k = 1$, we can define an approximation for $\Delta \mathbf{w}^i$, $\Delta \hat{\mathbf{w}}^i$, in terms of left and right singular unit vectors $\mathbf{X}^i$ and $\boldsymbol{\Delta}^i$ corresponding to the largest singular value, $\sigma^i$. The rank 1 approximation, which we call the principal eigenupdate of the batch, is then:

$$\Delta \hat{\mathbf{W}}^i \approx -\eta {\sigma^i}^2 \boldsymbol{\Delta}^i [\mathbf{X}^i]^T.$$

This represents the single best rank 1 approximation of the batch update, with $\eta$ the traditional learning rate. These values can be estimated over a streaming batch of size $n$ such that $\mathbf{X}^i \approx \frac{\mathbf{x}^{i,n}}{\|\mathbf{x}^{i,n}\|}$, $\boldsymbol{\Delta}^i \approx \frac{\boldsymbol{\Delta}^{i,n}}{\|\boldsymbol{\Delta}^{i,n}\|}$, and $\sigma^i \approx \sigma^{i,n}$ using the following update rules where $j$ runs from 1 to $n$:

$$\mathbf{X}^{i,j+1} = \frac{j}{j+1}\mathbf{X}^{i,j} + \frac{1}{j+1}\mathbf{x}^{i,j}\frac{(\boldsymbol{\delta}^{i,j} \cdot \boldsymbol{\Delta}^{i,j})}{\|\boldsymbol{\Delta}^{i,j}\|}$$

$$\mathbf{\Delta}^{i,j+1} = \frac{j}{j+1}\mathbf{\Delta}^{i,j} + \frac{1}{j+1}\mathbf{\delta}^{i,j}\frac{(\mathbf{x}^{i,j} \cdot \mathbf{X}^{i,j+1})}{\|\mathbf{X}^{i,j+1}\|}$$

$$\sigma^{i,j+1^2} = \frac{j}{j+1}\sigma^{i,j^2} + \frac{1}{j+1}\frac{(\mathbf{x}^{i,j} \cdot \mathbf{X}^{i,j+1})}{\|\mathbf{X}^{i,j+1}\|}\frac{(\mathbf{\delta}^{i,j} \cdot \mathbf{\Delta}^{i,j+1})}{\|\mathbf{\Delta}^{i,j+1}\|}$$

Afterwards the weight matrix is updated using the rank 1 estimators of the singular values. The next batch is calculated from the end condition of the previous batch such that $\mathbf{X}^{i+1,1} = \mathbf{X}^{i,n}$, $\mathbf{\Delta}^{i+1,1} = \mathbf{\Delta}^{i,n}$, and $\sigma^{i+1,1} = \sigma^{i,n}$. The previous best estimate is presumed to approximate the subsequent best estimate, which is true if the learning rate is sufficiently small[1].

This algorithm falls within a general family of noisy power iterations (M. Hardt and Price 2014), or power iterations performed on stochastic matrices, which are known to extract the eigenvectors of covariance matrixes. It is, additionally, a bi-iterative method for calculating both left and right eigenvectors (Clint and Jennings 1971; P. Strobach 1997).

Intuitively however, the algorithm can be interpreted as updating the "weighted average" activation and error based on the cross significance of its companion term. For example, the "estimated activation" of the layer, $\mathbf{X}^{i,j}$, is rotated significantly by $\mathbf{x}^{i,j}$ subject to the condition that $\frac{(\mathbf{\delta}^{i,j} \cdot \mathbf{\Delta}^{i,j})}{\|\mathbf{\Delta}^{i,j}\|}$ is large. If the error then of any particular input is small or pointing in an uncommon direction, the estimated activation does not change significantly. The same is true for $\mathbf{\Delta}^{i,j+1}$. This algorithm, in the context of estimating the SVD of a batch update matrix using streaming data, we call the streaming batch eigenupdate (SBE) algorithm.

An important feature of this approach is that it opens a tradeoff space between the software and the hardware. On one hand, it necessarily throws away a significant amount of the information from the batch, which results in a low rank approximation. Hence, for updates with higher rank, larger eigenvalue matrixes would be less well represented and therefore take a longer time to converge. On the other hand, this approximation, which is a form of compression, allows for a much more compact representation of the error, which has the potential to dramatically reduce hardware costs. One point to note here is that the smaller the rank of the weight update, the more representative a low rank approximation would be. Consequently, we might expect the eigenupdate to perform better for activation functions that lead to sparse updates, such as for rectifying activation functions like rectified linear units (ReLU).

Figure 1 shows an example of the potential effectiveness of our approach prior to running network models. It shows the relative significance of different singular values, subject to the normalizing condition $\sum_{p=1}^{r}\sigma_p^2 = 1$ for singular index value $p$ up to rank $r$. The plots show a representative matrix decomposition for a particular batch update in the middle of a conventional 728x256x32x10 network trained on MNIST to 90 % accuracy for the test set. Based on the relative magnitude of values for our example batch, ReLU activations can have as much as 60 % of the batch information contained in the first pair of singular vectors. From the cumulative contribution, we can see for sigmoidal activation, which squashes the outputs of the neurons, the first 10 pairs of singular vectors can capture as much as 95 % of the information contained within our example batch. We attribute this fact ultimately to the fact that despite the large sizes of matrixes in these networks, the complexity of the trajectories will ultimately be limited by significantly smaller number of classes which are used to train the networks. Others have observed that the trajectories of networks during training can be reduced to a smaller number of principal components (Antognini and Sohl-Dickstein 2018; E. Lorch 2016).

---

[1]Sporadically, one or the other singular vector becomes anti-parallel to the calculated true vector, causing either the other vector to also become anti-parallel or for the singular value squared, $\sigma^{i,j^2}$, to become negative. These erroneous sign changes always occur in pairs during the calculation of the approximate eigenupdate, so that the net sign is always correct.

### 3.2 Network Modeling and Experiments

For our experiments, we compare traditional approaches, stochastic gradient descent (SGD) and mini-batch gradient descent (MBGD), with our PCA based approaches, specifically doing the singular value decomposition (SVD) and the streaming batch eigenupdate (SBE) estimation of the batch between matrix updates. While MBGD and SVD cannot be efficiently implemented, SGD and SBE can. Figure 2 outlines the key distinctions in the process execution of the algorithm.

To compare these approaches, we choose a very simple network architecture of 728x100x10 neurons, using ReLU activation functions between layers and a cross-entropy loss function (Y. LeCun et al. 1998). To control for the fact that using batches reduces the overall number of updates per epoch, we use a learning rate optimizer prior to network simulations, which minimizes the loss for 5 epochs. There is a hard cutoff terminating our simulations after 900 epochs. Batch sizes were varied from $2^0$ to $2^{13}$. Networks are trained on the MNIST data set using the typical test-training partition. The exemplary series of networks trained below all began from the same randomly drawn starting condition.

### 4 Results

To illustrate the convergence of the SBE algorithm during the batch training process, we calculate the error, ε, for the converging the singular vectors, $\mathbf{X}^{i,j}$, to the true singular vectors, $\mathbf{X}^i$, as $\varepsilon = 1 - abs(\frac{\mathbf{X}^i \cdot \mathbf{X}^{i,j}}{\|\mathbf{X}^{i,j}\|})$, and similarly for the singular value as $\varepsilon = 1 - abs(\frac{\sigma^{i^2}}{\sigma^{i,j^2}})$.[1] Figure 3 shows convergence curves of these errors during network training for batch sizes 32 and 1024. While 1024 shows strong periodic behavior between updates and convergence of the singular vectors down to an accuracy below $10^{-3}$, the smaller batch size of 32 shows periodic behavior but no strong trends toward convergence of the approximate singular vector. Despite this weak convergence of the singular vector, the training of the network converges.

That the convergence of the singular vectors is not necessary to demonstrate convergence of the network makes sense because convergence of vectors during power iterations is often determined by the eigengap, or the gap between the target eigenvalue and the next smallest eigenvalue of a matrix (C. Musco and Musco 2015). A small eigengap leads to significant contamination of the target vector with other large eigenvalue vectors. This contamination complicates finding the eigenvector itself but still pushes a network to a lower value of the loss function.

Figure 4 shows that all the training algorithms reduce the training set loss function down to as low as $10^{-4}$. We find that reducing the training set loss down to $10^{-2}$ is sufficient to achieve 100 % accuracy on the training set and therefore about 97 % to 98 % accuracy on the test set. In these simulations, the SGD function is the fastest algorithm for training in terms of number of epochs, with MBGD, due to its parallelism, having significantly faster wall clock time. When re-plotting the data in terms of matrix updates, it's clear that the batch methods have an advantage in terms of minimizing the number of times the memory needs to be changed. However, these measures do not take into account the time that would be required to do the matrix updates in hardware. Since the SVD and SBE methods use only rank 1 updates, it takes less time for them to update the hardware by a factor of the number of elements in the crossbar.

These general trends can be seen in Figure 5, which shows the number of epochs and number of matrix updates needed to train the network to a training set loss of both $10^{-1}$ and $10^{-2}$. For this example, MBGD is clearly the highest performing on all metrics, decreasing the number of updates needed to train the network vs. SGD by more than two orders of magnitude at a batch size of 4096. For the SBD and SBE algorithms, the epochs to train grows much faster, and the number of matrix updates needed to train only falls by a factor of 20 compared to SGD and does so at a much smaller batch size of 128. For very small batch sizes, the SBE algorithm performs worse than the SVD algorithm, which we attribute to poor qualities of the update vector, but at higher batch sizes it outperforms the SVD algorithm, which we attribute to a mixture of better update quality but with added stochasticity lacking in the SVD approach due to the random degree of convergence and sampling of lower significance eigenupdates and singular vectors.

## 5 Discussion

For the example below, the SBE approach is lower performing than the MBGD approach in terms of number of epochs to train and number of matrix updates. However, its use would vastly accelerate the wall clock time of training in a hardware network since the transfer of the weights has the same complexity as the SGD approach, even in cases where the batches were stored in a local and parallel short-term memory array. Moreover, in the case of $k = 1$, calculating and storing the low rank versions of activations and error (left and right eigenvalues) take up significantly less area and compute ($O(a+b)$) as compared to the full rank ($O(a \times b)$) versions.

If a higher quality update were desired, the above algorithm could be extended to the calculation of multiple eigenupdates in parallel, similar to an Oja asymmetrical subspace network (E. Oja 1992). The application of $k$ eigenupdates would still be significantly faster than the time needed to transfer the point-wise or column-wise transfer for a full ranked batch update. Based on Figure 1, it's clear that a full rank transfer is unnecessary and possibly even detrimental if excess information leads to over fitting.

The critical challenge is determining the most efficient hardware implementation of the SBE algorithm. The major operations required are the summation, multiplication and division respectively. Among them the most computationally intensive part is the normalization operation, $\frac{\mathbf{x}^{i,n}}{\|\mathbf{x}^{i,n}\|}$. Since we may only be working with low precision, such as 4-bit precision, and only dealing with a linear number of computations vs problem size, the overhead of implementing these operations is significantly smaller when compared to their full rank counterparts. Digital implementations of such operations can be constructed with systolic array approaches, and if further energy efficiency is required, analog approaches can be used as well(Vanpoucke, Moonen, and Deprettere 1994).

An alternate analog approach which gets rid of the division operation altogether is borrowed from the original Taylor series formulation of the Oja equations, which replaces division with a multiplication and subtraction (See Supplemental). Such a calculation, though, may run into issues with numerical stability. However, the physical constraints of the system along with the parallel calculation of additional singular vectors could stabilize the algorithm. Calculating multiple singular vectors is known to accelerate convergence of the dominant vectors (Balcan et al. 2016). Moreover, future hardware could likely use short term memory cells, such as trench capacitors and FET's, to perform resistive multiplication and dot product operations in combination with Gilbert cells to scale the outputted values properly (Y. Li et al. 2018).

This is a rich tradeoff space which requires further exploration and is going to be part of follow-up work. Trading off the different system attributes, digital vs. analog, single vs. multi-rank, and batch size, can be used to build an optimal system for machine learning by efficiently calculating streaming batch eigenupdates.

## 6    Tables and figures

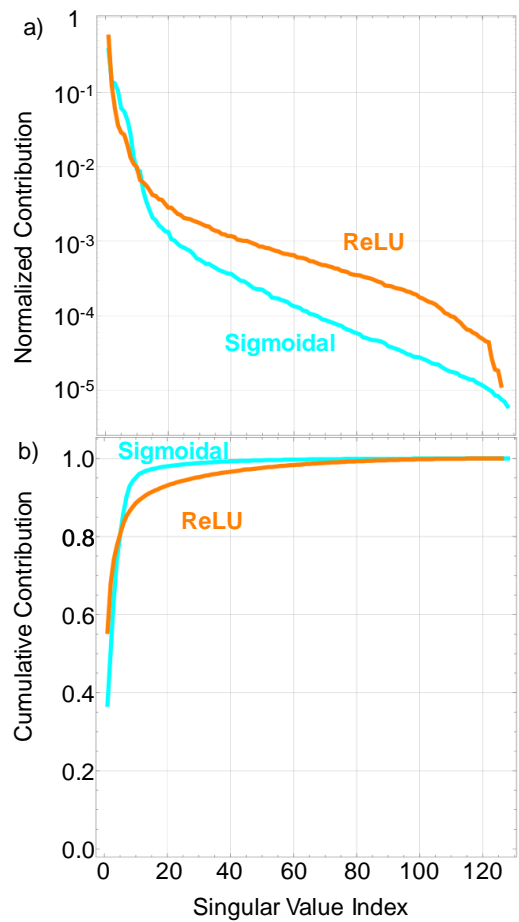

**Figure 1** a) Example of normalized singular values of the middle layer of a 728x256x128x10 network trained for MNIST with ReLU and sigmoidal activation.  The batch size is 10,000. b) Cumulative sum of the singular values. The sum of the first few vectors approaches the total sum, one, showing that they contain most of the batch information.

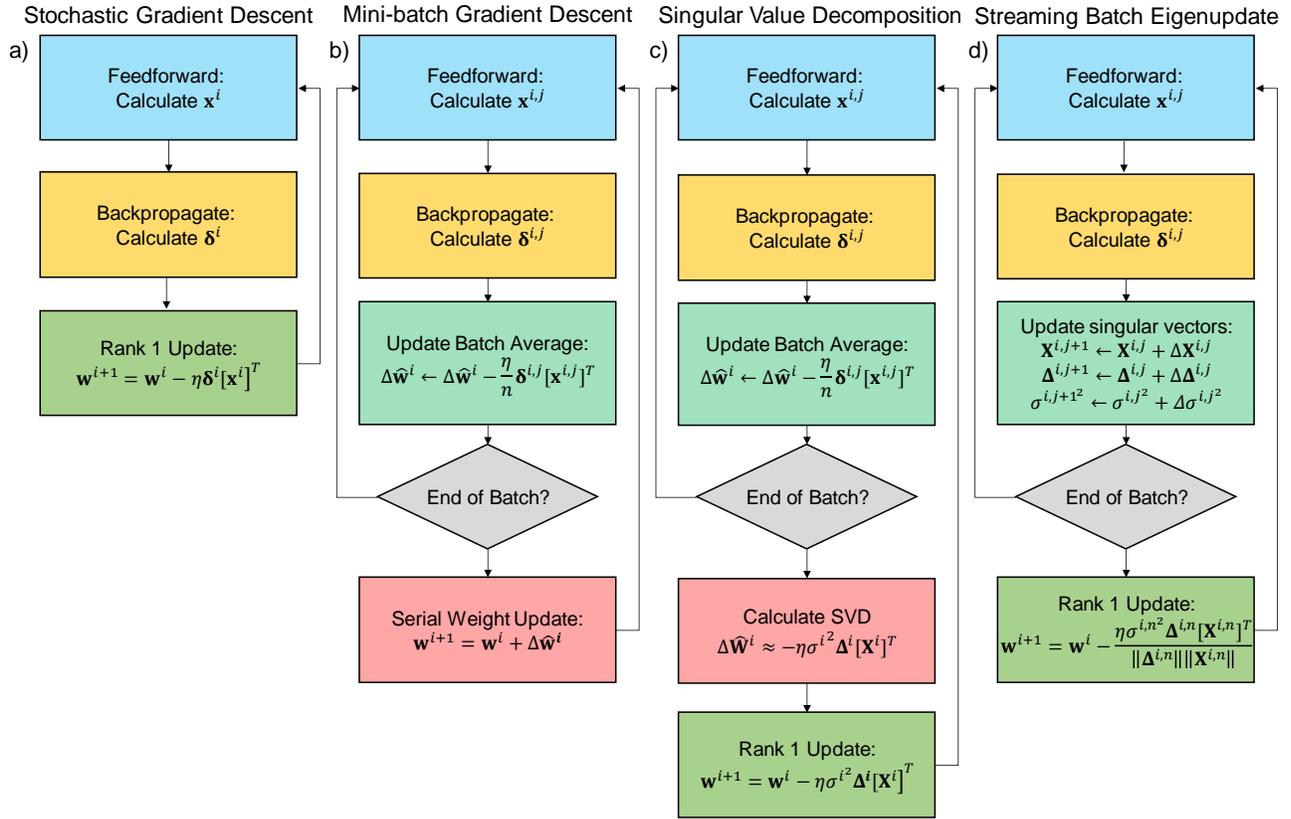

**Figure 2** Simplified comparison of the training algorithms for a) stochastic gradient descent (SGD), b) mini-batch gradient descent (MBGD), c) the singular value decomposition (SVD) approximation of the batch, and d) streaming batch eigenupdates (SBE). Both SGD and SBE are rank 1 and calculated on the fly, achieving the highest degree of acceleration.

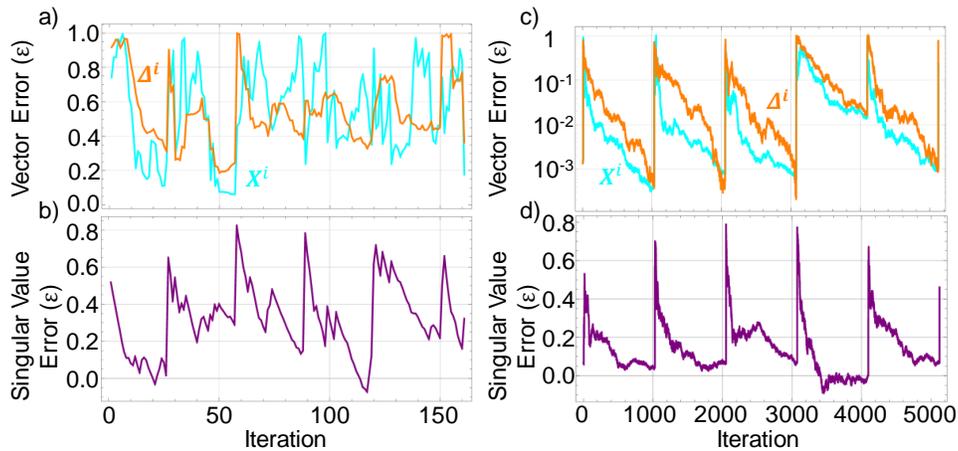

**Figure 3** Difference between SBE values and the full SVD values for a) and c) singular vectors $\mathbf{X}^i$ and $\mathbf{\Delta}^i$ as calculated by $\varepsilon = 1 - abs(\frac{\mathbf{X}^i \cdot \mathbf{X}^{i,j}}{\|\mathbf{X}^{i,j}\|})$ and b) and d) singular values as calculated by $\varepsilon = 1 -$

$abs(\frac{\sigma^{i^2}}{\sigma^{i,j^2}})$.[1] Batch sizes are 32 a) and b) and 1024 c) and d). The larger batches show greater fidelity with more iterations. The sharp increases in the difference correspond to the update of the weight matrix and subsequent change in the gradient.

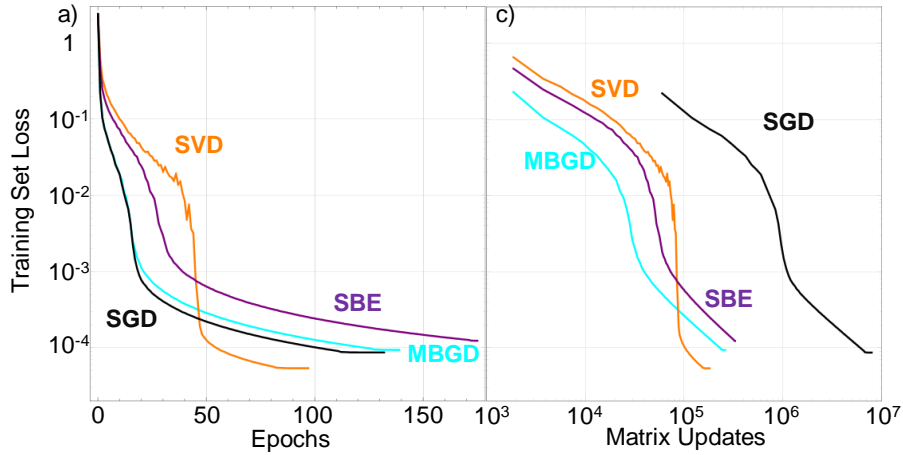

**Figure 4** Training set loss functions under different SGD and batch learning rules (batch size is 32) vs the number of a) epochs and b) matrix updates. The SVD and SBE algorithms required more epochs to train but fewer matrix updates.

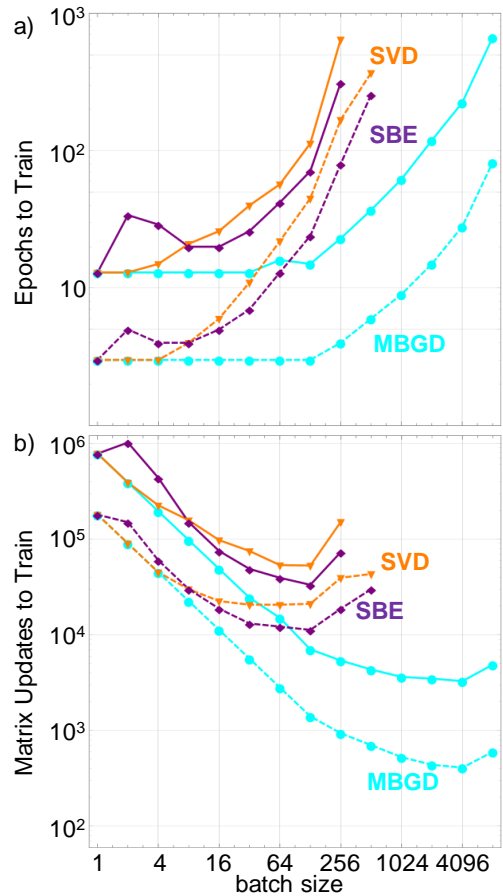

**Figure 5** Summary of impact of different training rules vs batch size including a) the number of epochs to train the training set loss function down to 0.1 (dashed lines) and 0.01 (solid lines), and b) the number of matrix updates to set the loss function to 0.1(dashed lines) and to 0.01 (solid lines). The SVD and SBE training rules increase the update efficiency, but not as much as full batch update.

## 7 Conflict of Interest

*The authors declare that the research was conducted in the absence of any commercial or financial relationships that could be construed as a potential conflict of interest.*

## 8 Author Contributions

BDH conceived of the streaming batch eigenupdate concept for training networks. BDH, MWD, and MDS developed the mathematical framework and algorithms implemented. BDH and MWD performed the network simulations. AM and BDH analysed implications of implementing the algorithm in hardware. All authors analysed the results and wrote the manuscript.

## 9 Funding

AM acknowledges support from the Cooperative Research Agreement between the University of Maryland and the National Institute of Standards and Technology Center for Nanoscale Science and Technology, Award 70NANB14H209, through the University of Maryland.

# Supplementary Material

## Taylor Expanded SBE Algorithm

If we assume that the SBE algorithm has been running for some time such that $\mathbf{X}^{i,j}$ and $\boldsymbol{\Delta}^{i,j}$ are already unit vectors and the learning rate is small (as could be in the case of large $j$, leading to a small learning rate $\xi$), we could formulate the algorithm first with an explicit normalization and then expand this in a Taylor series:

$$\mathbf{X}^{i,j+1} = \frac{\mathbf{X}^{i,j} + \xi \mathbf{x}^{i,j}\big(\boldsymbol{\delta}^{i,j} \cdot \boldsymbol{\Delta}^{i,j}\big)}{((\mathbf{X}^{i,j} + \xi \mathbf{x}^{i,j}(\boldsymbol{\delta}^{i,j} \cdot \boldsymbol{\Delta}^{i})) \cdot (\mathbf{X}^{i,j} + \xi \mathbf{x}^{i,j}(\boldsymbol{\delta}^{i,j} \cdot \boldsymbol{\Delta}^{i,j})))^{1/2}}$$

For small values of $\xi$, the expansion of the singular vector is:

$$\mathbf{X}^{i,j+1} = \mathbf{X}^{i,j}\big(1 - \xi\, \boldsymbol{\delta}^{i,j} \cdot \boldsymbol{\Delta}^{i,j}\, \mathbf{x}^{i,j} \cdot \mathbf{X}^{i,j}\big) + \xi \mathbf{x}^{i,j}(\boldsymbol{\delta}^{i,j} \cdot \boldsymbol{\Delta}^{i,j})$$

Which naturally implies the other values:

$$\boldsymbol{\Delta}^{i,j+1} = \boldsymbol{\Delta}^{i,j}\big(1 - \xi\, \boldsymbol{\delta}^{i,j} \cdot \boldsymbol{\Delta}^{i,j}\, \mathbf{x}^{i,j} \cdot \mathbf{X}^{i,j}\big) + \xi \boldsymbol{\delta}^{i,j}(\mathbf{x}^{i,j} \cdot \mathbf{X}^{i,j})$$

$$\sigma^{i,j+1^2} = \sigma^{i,j^2}(1-\xi) + \xi \; \boldsymbol{\delta}^{i,j} \cdot \boldsymbol{\Delta}^{i,j} \; \mathbf{x}^{i,j} \cdot \mathbf{X}^{i,j}$$

The necessary hardware values that need to be calculated and stored are the scalars $\mathbf{x}^{i,j} \cdot \mathbf{X}^{i,j}$, $\boldsymbol{\delta}^{i,j} \cdot \boldsymbol{\Delta}^{i,j}$, and $\sigma^{i,j^2}$ along with the necessary update vectors to the singular vectors. With suitable hardware, the MAC, multiplication, and subtraction operations can occur in parallel.